\documentclass[conference]{IEEEtran}
\usepackage{cite}

\usepackage[pdftex]{graphicx}
\usepackage[cmex10]{amsmath}
\usepackage{amsfonts}
\usepackage{array}

\usepackage{mdwmath}
\usepackage{mdwtab}
\usepackage{fixltx2e}

\usepackage{stfloats}

\usepackage{url}

\begin{document}
%
\title{Increasing Air Traffic: What is the Problem?}

\author{
\IEEEauthorblockN{Areski Hadjaz}
\IEEEauthorblockA{Thales Air Systems, Rungis, France\\
Email: areski.hadjaz@thalesgroup.com}
\and
\IEEEauthorblockN{Ga\'etan Marceau}
\IEEEauthorblockA{Thales Air Systems and TAO, Rungis, France\\
Email: gaetan.marceaucaron@thalesgroup.com}
\and
\IEEEauthorblockN{ \ \ \ \ \ \ \ \ \ \ \ \ \ \ Pierre Sav{\'e}ant}
\IEEEauthorblockA{ \ \ \ \ \ \ \ \ \ \ \ \ \ \ \ \ Thales Research \& Technology, Palaiseau, France\\
\ \ \ \ \ \ \ \ \ \ \ \ \ \ \ \ \ Email: pierre.saveant@thalesgroup.com}
\and
\IEEEauthorblockN{ \ \ \ \ \ \ \ \ \ \ \ \ \ \ Marc Schoenauer}
\IEEEauthorblockA{ \ \ \ \ \ \ \ \ \ \ \ \ \ \ TAO, INRIA Saclay, France\\
 \ \ \ \ \ \ \ \ \ \ \ \ \ \ Email: marc.schoenauer@inria.fr}
}




\maketitle

\begin{abstract}
Nowadays, huge efforts are made to modernize the air traffic management systems to cope with uncertainty, complexity and sub-optimality. An answer is to enhance the information sharing between the stakeholders. This paper introduces a framework that bridges the gap between air traffic management and air traffic control on the one hand, and bridges the gap between the ground, the approach and the en-route centers on the other hand. An original system is presented, that has three essential components: the trajectory models, the optimization process, and the monitoring process. The uncertainty of the trajectory is modeled with a Bayesian Network, where the nodes are associated to two types of random variables: the time of overflight on metering points of the airspace, and the traveling time of the routes linking these points. The resulting Bayesian Network covers the complete airspace, and Monte-Carlo simulations are done to estimate the probabilities of sector congestion and delays. On top of this trajectory model, an optimization process minimizes these probabilities by tuning the parameters of the Bayesian trajectory model related to overflight times on metering points. The last component is the monitoring process, that continuously updates the situation of the airspace, modifying the trajectories uncertainties according to actual positions of aircraft. After each update, a new optimal set of overflight times is computed, and can be communicated to the controllers as clearances for the aircraft pilots. The paper presents a formal specification of this global optimization problem, whose underlying rationale was derived with the help of air traffic controllers at Thales Air Systems. 
\end{abstract}



%
\IEEEpeerreviewmaketitle

\section{Introduction}
\label{sec:intro}
In the forthcoming years, an increasing demand on the worldwide airspace will result in restructuring the air traffic management systems. 
It will address the inherent problems of uncertainty, complexity and sub-optimality. 
A better exchange of information between the stakeholders would allow the enhancement of the current situation. 
Nevertheless, the information must be exploited in such a way that the complexity does not increase. 
Otherwise, the capability of the controllers to handle traffic will decrease, and this is the opposite of the initial goal. 
Therefore, decision support tools are essential for the data processing task in order to give only the relevant information to the human after a series of underlying analyzes which encompass many functionalities, such as prediction, correction and optimization. 
In addition, real-time monitoring is a core component that should be taken into account in the development of future systems.

In this work, we propose a model to address the uncertainty and perform optimization in the tactical phase of Air Traffic Flow Management. 
The novelty of this model is to directly address the uncertainty and the real-time monitoring aspects. 
The main benefit is to anticipate the congestion points by enhancing the information sharing between stakeholders. 
Also, this model could help improving the resilience of the system in case of weather hazards. 
We believe that an inter-sector coordination (Letter of Agreement) can be prepared through objectives on the trajectories. 
For the controller, it consists of bringing the flight to a geographical point within a time interval. 
This can be done by manually giving some clearances, or by communicating the objective directly to the pilot. 
Then, it can be easily entered in the Flight Management System (FMS) of the aircraft. 
These objectives, associated to some geographical points termed {\em metering points} in the following, are the core of our approach. 
In terms of acceptability, the responsibilities of the controllers do not change, because fulfilling the objectives will nevertheless be given a lower priority than the safety of the aircraft.

This article is organized as follows: Section \ref{sec:relatedWork} presents a literature survey of the air traffic flow management problems. 
Section \ref{section:frameworkDefinition} defines  the framework, while in Section IV the airspace model is presented together with a preliminary analysis of its properties. 
Section V describes the optimization problem over the model and Section VI fully describes a toy example. 
Section VII presents a Monte-Carlo approach in order to simulate trajectories over a realistic airspace. 
Finally, Section \ref{section:conclusion} concludes the paper and states some open questions that will be eventually addressed in our future research.

\section{Related Work}
\label{sec:relatedWork}
The literature on Air Traffic Flow Management is quite rich in the Operational Research community since the beginning of the 90s. 
Different problems have been addressed with different levels of complexity, and two different axes can be used to classify them. 
First, the static approaches consider a single stage while the dynamic approaches are multi-stage. 
Static problems perform an optimization once and for all, whereas dynamic problems construct partial solutions on-the-fly, based on regularly updated information, e.g. better forecast of weather and traffic demand. 
Second, the problems can be either deterministic or stochastic. 
In the stochastic case, the constraints and the decision variables are not known with certainty. 
Different scenarios are defined through a tree scheme, aiming at reducing the problem to an equivalent deterministic model. 
A thorough literature review of these approaches can be found in \cite{Agustin}, and the remaining of this Section will described the most well-known in turn.

The Ground Holding Problem minimizes the sum of airborne and ground delay costs when the demand of the runway exceeds the allowed capacity. 
It does so by assigning ground delays to flights. 
The first variant of the problem is the Single Airport Ground Holding Problem \cite{Odoni1987}. 
A stochastic and dynamic version of the same problem was later described \cite{Richetta1994}, and recently addressed in \cite{Mukherjee2007} by overcoming some limitations of the previous model such as modeling the change on marginal probabilities over a finite set of scenarios and allowing revisions to assigned ground delays of flights. In these works, the objective functions are a trade-off between efficiency and equity and might be non-linear.
The second variant is the Multi Airport Ground Holding Problem \cite{Vranas1994}. 
This model was the first one to model a network of interconnected airports and connecting flights with delays propagating through the network, and addresses the static and dynamic cases of the problem. 
However, strong assumptions are made: the sector capacities are unlimited and rerouting and speed changes are not allowed. 
These assumptions are unrealistic in a congested airspace such as the European one. 
This lead to introduce the Air Traffic Flow Management Problem with sector capacities and rerouting \cite{Bertsimas1998}. 
However, in this work, the decision variable of a time slice is set to true only if the flight enters a given sector exactly during time slice, resulting in tightening the structure of the linearized problem. 
Note that this work uses realistic instances with several thousand flights.
Finally, the Air Traffic Flow Management Rerouting Problem \cite{Bertsimas2011} was proposed, and is to-date the most complete description of the actual system. 
It integrates all phases of a flight, ground and air delays, rerouting, continued flights and cancellations. 
The validation instances are of the same size than the National Airspace of the United States. Also, \cite{Agustn2012a}, \cite{Agustn2012b} developed a formulation based on the routes of the network instead of the nodes, both for the deterministic and stochastic cases. 
All the approaches mentioned so far resulted in boolean problems that were solved with 0/1 programming techniques. 
These techniques are today powerful enough to address large-scale scenarios. 
A different type of problem setting lead to using stochastic optimization methods: the work in \cite{Oussedik1998} handles sector congestion with take-off delays and alternative routes while managing the airlines constraints.
In another approach, \cite{Barnier2001} modeled the slot allocation problem, which consists of assigning slots to flights in the sector and respecting the capacity constraint. 
The resulting problem was solved with Constraint Programming. 
Furthermore, the minimization of air traffic complexity in a multi-sector planning paradigm was addressed in \cite{Flener2007}, also with Constraint Programming. 
Moreover, optimal path planning under weather uncertainty is addressed by \cite{Aspremont2006} with Dynamic Programming techniques.

In all previous models, uncertainty was eventually addressed by using a finite number of scenarios. 
Nevertheless, the sector load prediction is an essential issue to consider. 
As a matter of fact, a poor estimation of the time required to travel along a route will generate unnecessary regulations. 
Approaches based on trajectories have failed to predict the state of the airspace beyond a time horizon of 20 minutes, leading to the design of aggregate models \cite{Sridhar2008a}. 
These models have been used by \cite{Sun2006} to achieve control over one sector with a multicommodity network. 
Due to the fine time-discretization along the edges, the size of the optimization problem is huge. 
However, to the best of our knowledge, no system has yet addressed the real-time monitoring and optimization of a large-scale network by modeling the flight plan with temporal uncertainty in a continuous domain. The purpose of this paper is to propose a research path toward such a system, sketching both the formal and the numerical aspects of the resulting problem.

\section{Framework Definition}
\label{section:frameworkDefinition}
The framework used in this work is different from the ones of the literature. 
The time horizon considered is from current time up to 2 hours which is referred to the strategic phase in air traffic control. 
The idea behind the strategic phase is, on the one hand, to bridge the gap between the air traffic flow management (ATFM) and the air traffic control (ATC), and on the other hand, to bridge the gap between the ground, the approach and the en route phases. 
This is named a gate-to-gate solution. 
The ATFM is responsible for validating the flight plans of the entire day by considering the sectors capacity constraints while the ATC is responsible to ensure the aircraft separation and to minimize the delays. 
The idea behind the strategic phase is to take into account the information from the ATC and predict accurately the future sector load. 
Then, regulations or clearances can be applied on incoming flights in order to respect the sector capacity and reduce the complexity at the ATC level. 
In other words, the current information is used to reduce the future complexity. 
Today, the controllers are responsible for applying regulations or clearances to manage their airspace. 
As said previously, the decision support tool must not increase the workload by communicating too much information. 
We think that objectives on metering points will suffice to create the multi-center collaboration. 
So, the responsibilities of the controllers are not impacted, but they have additional information on the global state of the airspace i.e. a low value on temporal objective indicates that the flight can go directly to the next metering point without causing a congested situation in the following sectors. 
On the other hand, a high value indicates that airborne delays should be applied to reduce the workload in the following sectors. 
From the ATFM view, the model integrates the Central Flow Management Unit (CFMU) slot allocations of the European context. 
It is important to notice that the uncertainty comes essentially from these departure slots. 
A slot is defined around a calculated time of take-off $t_d$ as an interval: $\left[ t_d-5, t_d+10 \right]$. 
So, the uncertainty of departures is around 15 minutes. 
The following model will be a way to evaluate the impact of such intervals.
The objectives are affected by the evolution of the uncertainty generated by weather conditions, aircraft performances and interaction between aircraft. 
In a purely deterministic world, a global optimization should be done once and for all in order to find the global optimal plan. 
As seen, such problems can be solved for huge airspace, but the complete solution can hardly be implemented due to the departure slot uncertainty of 15 minutes. 
Here, this uncertainty will be integrated at the model level, and monitored in a closed-loop way, as it diminishes with time. 
If everything goes as planned, the objectives will be slightly modified according to the evolution of the variance. 
In the case that an unforeseen event occurs, the objectives will be modified in function of its severity.

\section{System Overview}
\label{sec:systemOverview}
This Section introduces the rationale of the proposed way to manage the uncertainty and optimize the state of the airspace at different time scales. 
The temporal scope is from the current time up to 2 hours. 
The geographical scope is in the order of multiple centers in a congested area, e.g., the European airspace.
The system consists of three processes: model creation, monitoring and optimization. 
The model creation receives new flight plans, and creates one Bayesian Network for each flight plan once and for all (in particular, all possible alternative routes are known at creation time). 
The periodic monitoring adjusts the parameters of the model following observations from the real world. 
This process ensures that the model of the actual situation is consistent with reality, a critical issue for doing accurate prediction when sampling. 
The optimization process optimizes the current situation by proposing modifications of the flight plans in order to reach an optimal state. 
Then, these new plans (schedules at different metering points) are communicated to the controllers, who implement them their own way.

\subsection{Inputs}
The inputs of the system are the airspace structure and the flight plans. 
The airspace is defined by a set of metering points located in different sectors where ingoing and outgoing points are distinguished from the others. 
One flight plan is defined as a directed acyclic graph (DAG) where the nodes are defined as pairs (metering point, schedule), containing all possible alternative routes for this flight. 
A flight plan also includes the estimated time of departure, the estimated time of arrival and a statistic of the time required to travel along the routes linking the successive metering points. 
This statistic can be determined with flight position recordings when they are sufficient or any model of trajectory uncertainty. 
It can also be defined with a simulator that can model non-deterministic phenomenon like the effect of the wind, errors on the cruise speed, on the flight altitude changes etc. 
In any case, the following model is enough general to take into account any statistic under the form of a probability density function. 
To obtain it, one can generate the histograms and approximate them with density estimation techniques., e.g. kernel density estimation. 
To reflect the aircraft performance, a lower bound and an upper bound are also associated to routes.

\subsection{Trajectory Model}
The model used to express the uncertainty around a trajectory is a Bayesian Network (BN). 
This is a natural choice since it is a DAG composed of random variables (RV) and their conditional dependencies. 
A node is equivalent to a RV and an edge between two nodes represents that the first RV has a direct influence on the second one. 
The BN framework can then be used to do inference when new information arrives as time goes by. 
For more information on BN, see \cite{Koller+Friedman:09}. 
Compared to other approaches from the literature, BN is a framework that deals directly with uncertainty. 
Some continuous distributions can express a rich set of scenarios, potentially an infinite number, with only few parameters. 
It is mainly used to infer unobserved variables and so, it is well-adapted for predicting the time of arrival on the following metering points. 
Many algorithms exist to do inference. 
Also, learning is an important topic in BNs. 
Techniques exist for learning the parameters of the RV. 
This will be important in the monitoring process. 
Optimization can be easily integrated in this framework.
Note that exact algorithms, like message passing, efficiently use the structure of the network to do this task. 
In the case of huge networks, powerful approximate inference algorithm, like particle-based, exists too. 
In our case, this set of tools is well adapted to the graph structure of the airspace network. 
The interactions between metering points, flights and sectors can be captured through this representation.
In the context of ATFM, the metering points are 3D points of interest (longitude, latitude, altitude) such as coordination points between sectors or convergent flows points. 
In order to keep the representation as compact as possible, only the points included in the flight plan at hand are part of the network. 
Let $G_f=(V_f,E_f)$  be the DAG giving the anticipated routes as well as acceptable alternative routes of flight $f$.
Two particular vertices of the DAG are distinguished: $V_0$ is the origin and $V_d$ is the destination. 
Associated to a point i of the route, a probability density function (PDF) $p_{T_{f,i}}: \mathbb{R} \rightarrow [0,1]$ gives the probability distribution that $f$ will flight over this point over time. 
At this point, the PDF of $V_0$ is known, i.e. the CFMU slot or the estimated time of entrance in the airspace. 
Let's determine the PDF of the others. 
To this end, we use the PDF of $T_{f,i \rightarrow i+1}$, the probability distribution of the time required to go from $i$ to $i+1$. 
Also, we make the assumption that the time required to go from $i$ to $i+1$ is independent of the time of arrival at $i$. 
Then, we suppose that $T_{f,i+1}=T_{f,i}+T_{f,i \rightarrow i+1}$, i.e. the time at the current point is the sum of the time at the previous point with the time required to go from the previous point to the current one. 
To make sense, the PDF of $T_{f,i+1}$ will be determined in function of the PDF of $T_{f,i}$ and $T_{f,i \rightarrow i+1}$. 
We use the joint probability distribution $p_{T_{f,i},T_{f,i \rightarrow i+1} }$ and we integrate along the line $t_{f,i \rightarrow i+1}=t_{f,i+1}-t_{f,i}$. 
In terms of density probability function, we have:

\noindent
$\displaystyle
p_{T_{f,i+1}} (t_{f,i+1}) 
$
\begin{eqnarray*}
  &=& \int_{- \infty}^{\infty} p_{T_{f,i},T_{f,i \rightarrow i+1}} \left( t_{f,i},t_{f,i+1}-t_{f,i} \right)  dt \\
  &=& \int_{- \infty}^{\infty} p_{T_{f,i}}(t_{f,i}) p_{T_{f,i \rightarrow i+1}} \left(t_{f,i+1}-t_{f,i} \right)  dt \\
  &=& \left[ p_{T_{f,i}}*p_{T_{f,i \rightarrow i+1}}  \right] (t_{f,i+1} )
\end{eqnarray*}

The second line is obtained with the independence assumption.
The third line expresses the fact that the PDF of the sum of two independent RVs is the convolution of their PDFs.
If we suppose that the trajectory is defined as a series of RV $\left( T_{f,i},T_{f,i \rightarrow i+1} \right)_{i \in \left\{ 0..n-1 \right\}}$ where $n$ is the number of points of the trajectory, we can find that the PDF of any RV $T_{f,i}$ is the convolution of all the previous PDF in the sequence. 
This can be proven by induction.
Then, in a real-time system, we would like to take into account new information. 
This can be done with the conditional PDF:
\begin{equation}
\label{eq:cond}
p_{T_{f,i+1} |T_{f,i}} (t_{f,i+1} | t_{f,i} ) = p_{T_{f,i \rightarrow i+1}} ( t_{f,i+1}-t_{f,i} )
\end{equation}
which simply translates the PDF of the time of traveling of $t_{f,i}$ time unit. 
This does not change the previous result. 
As a matter of fact, we can remove the PDF of the observed random variables, because there is no more uncertainty associated to them, and replace the first remaining PDF with a conditional PDF.
At this point, we have the necessary concepts to create the trajectory structure in our BN. 
This corresponds to a path, the sequence of metering points, and the RVs of traveling linked to them. 
So, we can make a query on the BN by using the joint PDF and its relationship with conditional PDF. 
For the sake of notation, let $\Pr ( T_{f,i} \in dt ) = p_{T_{f,i}}(t_{f,i}) dt$ and $\Pr (T_{f,i} \in dt|T_{f,j} \in dt) = p_{T_{f,i} |T{f,j}}(t_{f,i},t_{f,j})dt \; dt$.
Here, we use the Markov property:
\begin{equation*}
\Pr(T_{f,i+1}, \dots, T_{f,i}) = \Pr(T_{f,i+1} |T_{f,i} )
\end{equation*}
which states that the conditional pdf of the future point does not depend on the past points given the current one. 
This simplifies the computation of the queries that required the joint PDF. 
With the chain rule and the Markov property, one can write:
\begin{eqnarray*}
\Pr (\cap_{i=1}^n T_{f,i}) &=& \Pr(T_{f,1}) \prod_{i=2}^n \Pr (T_{f,i} | \cap_{j=1}^{i-1} T_{f,j}) \\
                        &=&  \Pr(T_{f,1}) \prod_{i=2}^n \Pr (T_{f,i} |T_{f,i-1} )\\
                        &=&  \Pr(T_{f,1}) \prod_{i=2`}^n \Pr (T_{f,i \rightarrow i+1}) 
\end{eqnarray*}
where $\Pr (T_{f,i \rightarrow i+1})$ is the translated PDF defined at Eq.\ref{eq:cond}.
Notice that, for now, the graph structure of the resulting BN is a forest, a disjoint union of trees. 
Consequently, all trajectories are independent.
To manage the case of diversions, we change directly the structure of the BN by removing the old path and adding the new one. 
Notice that after the modification, all RVs connected by a path to any modified RV are impacted and the associated PDFs must be computed.
The modeling of the uncertainty of the trajectory will be useful during the optimization process to evaluate the objective function and the constraints.
This should also capture the actual situation of the airspace updated by the monitoring process. 
Contrary to the existing works on stochastic optimization of the ATFMP, the uncertainty is modeled directly in the trajectory through the metering points.

\subsubsection{Sector Occupancy Model}
In the ATFM context, the usual way to measure the complexity of an airspace configuration is to count the number of flights that will go through the sector during a given time interval (e.g., one hour). 
Given the flight plans of the day, the complexity of every sector is predicted in order to determine the potential congestion time, and when appropriate, to issue regulations for certain flights. 
To this end, a capacity threshold on the number of flights is used. 
Similarly, in the proposed model, the sector constraint enforces that the number of flights in a sector is below a threshold during a given time slice. 
Nowadays, the time slice has a constant size of one hour. 
The reason behind this coarse discretization consists to absorb the uncertainty of the trajectories. 
In other word, we can be sure that the flight will be in that interval at a given moment. 
This has the effect to lower the effective capability of the controller to handle traffic.
As an example, assuming that the threshold of the capacity constraint is two flights and the time slice is 60 minutes, a sector would be congested whereas two flights could land at the beginning and one flight could take off at the end of the time slice. 
With a finer discretization of 15 minutes, only the first time slice will be congested. 
Nevertheless, to achieve a finer discretization, the prediction capabilities of the model must be of the same order of magnitude than the time slice size. 
This will be an important element to validate on the real dataset.
To define formally the sector constraint, let $S_{i,\left[ t_0,t_1 \right] }$ be the random variable that models the number of flight traveling in sector $i$ during the time interval $\left[ t_0,t_1 \right]$. 
In the BN, we create a node $S_i$ connected to every boundary points for every trajectories. 
Then, the PDF associated to $S_i$ is defined in function of PDF of the boundary points, which can become easily cumbersome.
Notice that to be a valid PDF, this definition shall respect the following property:
\begin{equation*}
\label{eq:sumToOne}
\sum_{j=0}^{\infty} \Pr (S_{i,[t_0,t_1]}=j) =1, \quad t_0<t_1 \in \mathbb{R} \cup [-\infty,\infty] 
\end{equation*}
i.e. that the probability that there are any number of aircraft at any time interval is equal to 1. 
Notice that this formulation can be used to describe the sector occupancy for any intervals. 
It simply changes the integration bounds of the underlying PDFs. 
Finally, let $C_{i,[t_0,t_1 ] }$ be the Bernouilli RV that models the fact that a sector $i$ is congested during the time interval $\left[ t_0,t_1 \right]$.
Then, $\Pr (C_{i,[t_0,t_1 ]} =1)=P(S_{i,[t_0,t_1]} > c_i )$ where $c_i$ is the capacity of the sector $i$.

\subsection{Monitoring}
Finally, the monitoring process is responsible for maintaining a consistent model of the actual situation of the airspace. 
Information from radars, aircraft and the weather centers can be taken into account to estimate the values of the distributions of the random variables of the BN.
The task undertaken in this process is filtering which estimates the current value given past observations.
Practically, the monitoring changes the PDFs and integrates observations of overflight time by creating conditional PDFs.
By example, when a flight takes off, the exact time of departure in known and the 15 minutes of uncertainty vanishes.
This must be taken into account in the subsequent optimization phases.

\section{Optimization}
Now that we have an airspace model, we want to optimize the time of over flight in order to reduce the probability of congestion for every sector and minimize the delays incurred by resulting regulations. 
To do so, we need to define an optimization problem with an objective function to minimize, decision variables and constraints.

\subsection{Objective function}
For the purpose of optimization, objective functions have to be defined. 
As a matter of fact, the interests of each stakeholder of the airspace are often antagonistic or ill defined.
A cost index quantifies the benefit of a trajectory for an airline.
Usually, this cost index is not communicated explicitly to the controllers and so, integrating the airlines preferences in the model can be difficult at this point.
Nevertheless, the initial flight plan is supposed to reflect the interest of the airlines and is validated by the CFMU.
Let $A$ be the vector of time of arrival for every flight, usually given by the airline.
With our airspace model, we can obtain an estimator of this value with the expected time of arrival:
\begin{equation*}
\hat A_f = \mathbb{E}(T_{f,a} )= \int_{-\infty}^{\infty} tf_{T_{f,a}}(t)dt
\end{equation*}
Consequently, the cost function shall penalize the delays on the expected time of arrivals:
\begin{equation*}
\mathcal{F}(\hat A,A) = \sum_{i=1}^N \left| \hat A_i - A_i \right|^p \quad p>1
\end{equation*}
where $\hat A$  is the vector of expected time of arrival for all the flights. 
The definition of the aggregation of the cost function of individual flights to obtain a global cost function is not unanimously accepted. 
As a matter of fact, it can have a major impact on the benefits of the airlines. 
The equity is a serious issue in the domain and must be addressed in this model. 
To do so, we use a super linear function \cite{Bertsimas2011}, where the exponent $p$ will penalize the situation where all the regulations are assigned to a few flights for the benefit of the others. 
Regarding the fitness landscape of the objective function defined previously; it is indeed possible that it is multimodal. 
Consider for instance two flights that wish to enter a sector that has a capacity threshold of one. 
Two optimal solutions are to delay either the first or the second flight. 
This kind of symmetry implies that the landscape is multimodal.

\subsection{Decision Variables}
The decision variables model two types of regulation. 
The first one is the rerouting regulation, which modifies the path of the original flight plan. 
Here, we simply choose to keep the initial route or to choose an alternative in the set of possible routes for a given aircraft. 
Once an alternative is chosen, the BN must be modified and inference must be recomputed. 
This operation has a computational cost and an operational cost in terms of complexity. 
For this reason, the modification of these decision variables shall be penalized. 
Nevertheless, when observing the change in the cost function with different alternatives, we can assess of the robustness of the solution with rerouting. 
Moreover, it can be a mean to evaluate if a rerouting strategy is appropriate to decrease the cost function of a flight. 
Besides, this permits to model the uncertainty around a weather phenomenon that will require the rerouting of a flight.
Once the rerouting decision variables are fixed, the second regulation concerns the times of arrival on the metering points. 
This is done through the parameters of the chosen traveling time PDFs of the airspace model. 
On one hand, a change on the mean can suggest that the flight might change its current speed and, on the other hand, a change on the variance might suggest that the flight will commit to arrive at the metering points with a greater precision. 
Besides, these parameters must be bound to respect the aircraft performance constraints. 
An upper bound for the number of decision variables is $A \cdot M \cdot P$ where $A$ is the cardinal of the set of alternatives, $M$ is the maximum number of traveling time RV per alternative and $P$ is the maximum number of parameters for modifying the PDFs of traveling time. 
Consequently, the number of decision variables can rapidly increase in function of these values.

\begin{figure}[tbh!]
\centering
\includegraphics[width=2.5in]{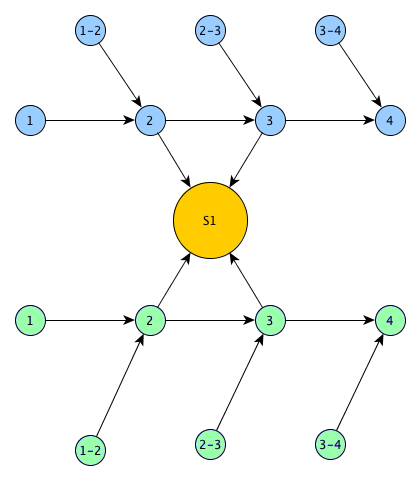}
\caption{BN for two trajectories and a sector}
\label{fig:ex}
\end{figure}

\subsection{Sector Constraint}
The sector constraint can be used as a hard constraint or a soft constraint. 
For the first case, a feasible solution must satisfies $\Pr (C_{i,[t_0,t_1 ]}>c_i ) \leq \epsilon$, i.e. that the probability that the sector $i$ is congested during the time interval $[t_0,t_1 ]$ is under a threshold $\epsilon$. 
For the second case, the soft constraint will be part of the objective function in order to minimize the probability that a sector is congested.

\section{Example}
In this section, we present a toy example of use of the airspace model. 
This will permit to understand the limits of a theoretical approach with very simple PDFs and will translate by the use of approximation methods, which are simpler for computer-based simulations.
So, let's imagine the case of two aircraft following the same flight plan. 
The sequence of waypoints is: 1-2-3-4. 
2 is an ingoing waypoint and 3 is an outgoing waypoint of the sector S1. 
Table 1 gives the parameters of the uniform distribution of the input PDFs.   
The expected time of arrival for both flights is at time 46. 
Also, we choose a capacity of 1 aircraft for sector S1. 
Figure \ref{fig:ex} shows the resulting BN where the blue nodes are RVs of the trajectories and the big yellow node is the RV of the sector occupancy.
 
With the d-separation in BN, we can see that the two trajectories are independent if S1 is unknown because it is a v-structure. 
If S1 is given, then the information on one trajectory will impact the PDFs of the other trajectories. 
Moreover, we can see the Markov property because there is only one edge going out of a RV of metering point going toward another metering point.

\begin{table}[h]
\centering
\caption{PDFs given as input}
\begin{tabular}{|c|c|c|}
\hline
&Lower Bound&Upper Bound\\ \hline
1&	-5&	10\\
1-2&	10&	12\\
2-3&	15&	20\\
3-4&	12&	18\\ \hline
\end{tabular}
\label{tab:instance}
\end{table}

Now, imagine that the PDF at the point 1 and the PDFs of the traveling time between points are all uniform distribution: $\mathcal{U}(t_0,t_1)$. 
The PDF at point 2 is then the convolution of two uniform distributions. 
The PDFs at points 3 and 4 are the convolution of the resultant and a uniform distribution.

\begin{figure*}[!t]
  \centerline{
    \begin{tabular}{c}
      \begin{tabular}{ccc}
        \includegraphics[width=2.5in]{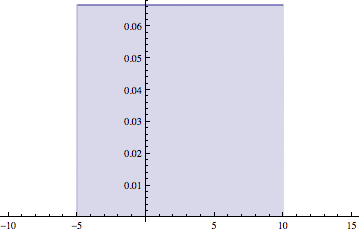} & ~~~~~ &
        \includegraphics[width=2.5in]{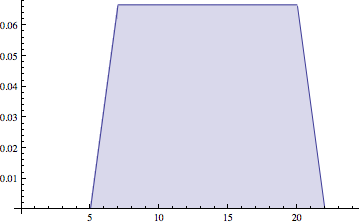} \\
        Prior on departure point & & Resulting PDF on point 2
      \end{tabular}
      \\~\\
      \begin{tabular}{ccc}
        \includegraphics[width=2.5in]{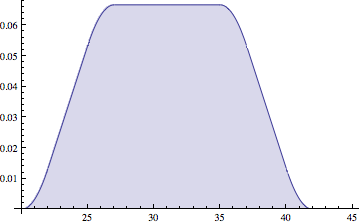} & ~~~~~ &
        \includegraphics[width=2.5in]{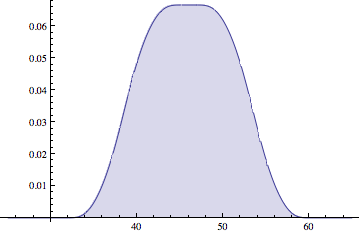} \\
        Resulting PDF on point 3 & & Resulting PDF on point 4
      \end{tabular}
    \end{tabular}
  }
  \caption{Convolution resultants on PDFs}
  \label{fig:convolution}
\end{figure*}

Figure \ref{fig:convolution} presents the resulting PDFs obtained by the convolution operator. 
The up left figure shows the PDF of the origin point 1 where we choose the CFMU interval. 
The up right figure shows the PDF at point 2, the down left shows the PDF at point 3 and the down right figure shows the PDF at point 4.
We can see that the number of pieces to define the function increases rapidly. 
For the PDF of 4, the function is defined with 24 pieces.
Now, the probability, modeled with a Bernouilli distribution, that the flight $i$ is not in sector 1 in the time interval $[t_0,t_1]$ is:\\

\noindent
$\displaystyle
\Pr(\overline{F_{i,s,[t_0,t_1]}})$ 
\begin{eqnarray*}
&=&\Pr(T_{i,2}>t_1) + \Pr(T_{i,3} <t_0 ) - \Pr(T_{i,2} > t_1,T_{i,3} < t_0 )\\
&=&\Pr(T_{i,2}>t_1) + \Pr(T_{i,3} <t_0 ) \\
&=&\int_{t_1}^{\infty} p_{T_{i,2}}(t) dt + \int_{-\infty}^{t_0} p_{T_{i,3}}(t) dt
\end{eqnarray*}

which corresponds to enter the sector after or to exit the sector before the interval. 
Note that these two events are mutually exclusive. 
Consequently, the probability to be in the sector is: $P(F_{i,s,[t_0,t_1 ]})=1-P(\overline{F_{i,s,[t_0,t_1 ]}})$. 
So, we can define the probability that the sector 1 will be congested:\\

$\displaystyle
\Pr (C_{1,[t_0,t_1 ]}) $
\begin{eqnarray*}
&=& \Pr (S_{1,[t_0,t_1 ]} >1)\\
&=& \Pr (S_{1,[t_0,t_1]}=2)\\
&=& \left[ \Pr (F_{1,1,[t_0,t_1 ]}) \right] \left[ \Pr( F_{2,1,[t_0,t_1 ]} ) \right]
\end{eqnarray*}

From the last equality, we remark that the flights are not required to be at the same moment in the sector to be taken into account in the probability. 
Consequently, in this case, the probability that the sector is congested during $[- \infty, \infty ]$ is equal to one. 
For the same considerations as described in the sector occupancy model section, it is important to consider small intervals. 
On the contrary, the intervals must not be too small because the probability that the two aircraft will be in the same sector during a small interval decays rapidly.
As an example, let $t_0=10$, $t_1=20$ and $\epsilon = 0.75$, we obtain $P(C_{1,[10,20]})=\frac{196}{225} \approx 0.87 > \epsilon$ and so, we consider that a regulation must be undertaken in order to reduce this high probability. 
For now, the objective function is equals to 0 since the expected value of the PDF at point 4 is 46. 
If we change the parameters of the uniform distribution of 1-2 for $[12,14]$, that is we delay a flight for 2 minutes, then the probability drops at $\frac{56}{75} \approx 0.747 < \epsilon$. 
On the other hand, the objective function is now equal to 2, for $p=1$. 
This toy example shows that the model is adequate to model the uncertainty of the trajectory at a high-level. 
Besides, we can see that the objective function does not take into account the variance of the RV of arrival time. 
In the latter, this issue shall be addressed.

\section{Monte-Carlo Approach}
\label{sec:MonteCarloApproach}
From the example, we can see that working with the PDFs with a symbolic computation approach can rapidly become cumbersome. 
Consequently, we will rely on Monte-Carlo simulations to estimate the sector congestion PDFs. 
From \cite{AndrieuFDJ03}, we know that a Monte-Carlo method used the fact that:
\begin{equation*}
\frac{1}{N} \sum_{i=1}^N f( x^{(i)} ) \xrightarrow{a.s.  N \rightarrow \infty} \int_X f(x)p(x) dx
\end{equation*}
where $x^{(i)}$ is a sample drawn from the PDF $p(x)$. 
This is an unbiased estimator and, by the law of large numbers, it will converge almost surely to the expected value. 
In our case, scenarios are built with trajectory sampling for statistical analysis on expected time of arrival and sector occupancy. 
A scenario is complete when it gathers one sampled trajectory per flight plan. 
This process gives as an output a set of $M$ scenarios $D={\psi[1], \dots,\psi[M]}$. 
The elements of this set are referred to particles in an approximate inference context. 
Of course, the accuracy of the Monte Carlo simulations depends directly on M and a sensitivity analysis should be done on the real dataset in order to determine the order of magnitude of M that is mandatory to ensure a good approximation.

\subsection{Stochastic Optimization Problem}
The goal of modeling a Stochastic Optimization Problem is to address uncertainty over the outcomes of a system by considering the most likely ones. 
In our system, this is done with Monte-Carlo simulation, which generates the scenario set D. 
The optimization process contains its own copy of the airspace model for evaluating the solutions without affecting the monitored airspace model. 
Then, the optimization process starts from the actual situation as a default solution and generates a new set of parameters for the model. 
A trajectory sampling method is used to generate a new set of scenarios and these are statistically analyzed in order to determine the variation of the aggregate cost function under the sector capacity. 
A possible stopping criterion for this iterative process is to monitor the value of the cost function, and to stop whenever it stays stable during a predefined time interval. 
This stopping criterion should work even in the case where we do not have enough information on the landscape of the cost function. 
However, other stopping criteria could be envisioned. 

\subsection{Sector Constraint Approximation}
As seen previously, an estimate of the probability of congestion can be the ratio between the numbers of scenarios where the sector is not congested over the total number of scenarios. 
This probability is computed for every time slice and so, this creates a stochastic process. 
So, a finer discretization has a cost in the number of random variables in this stochastic process to be estimated.
An efficient way to estimate this stochastic process is to determine the time intervals during which the flight is in a sector. 
This is straightforward with the sampled times at the entrance and exit points of the sector. 
Afterward, we can count the number of intervals that intersect with the time slice and compare it to the threshold.

\subsection{Objective Function Approximation}
In the Monte-Carlo approach, the landscape of the objective function is not explored directly, but approximated with a huge number of simulations. 
In this case, it is often more relevant to maintain a pool of good solutions since a solution is better than another solely with a given probability.
Finally, regularization terms will probably be necessary in the objective function, in order to obtain realistic solutions. 
The rerouting must be penalized since it adds a workload to the controller and the pilot. 
The means of the traveling RVs should also be modified only slightly, and in a consistent way, in order to minimize the number of changes in the flight behavior. 
Finally, the variance should always reflect the uncertainty of the system and should be integrated in the objective function. 
These considerations will be more easily taken into account with the Monte-Carlo approach due to its flexibility.
More generally, it is very likely that several different objectives will have to be considered for realism. 
Most probably, the optimization problem might become multi-objective - even though the ultimate output of the optimization process in the context of the proposed system should be a single set of changes for all flights. 
Hence some decision-making algorithm might also be needed here - unless some human supervision can be imagined. 
This could lead to the creation of a new role, e.g. multi-sector planner, or reinforce existing roles as the flow manager who will be in charge of using the system.

\section{Conclusion and Perspectives}
\label{section:conclusion}
A new system has been presented, in order to cope with uncertainty, complexity and sub-optimality. 
The novelty of this system is to define the trajectories as stochastic objects, i.e. parts of a Bayesian Network. 
Thereafter, these objects evolve with the real system and are updated by a monitoring process to keep a consistent model of the actual situation. 
From this model, an optimization process determines the optimal airspace configuration towards which the system should lean. 
The optimal airspace is communicated to the controllers in the form of objectives on the trajectories. 

This model is compatible with the current Air Traffic Control system and does not imply a complete reorganization of control centers. 
This approach prepares future enhancement based on the work done in SESAR and NEXTGEN projects by facilitating 4D trajectory integration in order to minimize the uncertainties, and thus increase the system relevance.
As a matter of fact, the controller is still responsible for rerouting flights at the tactical level. 
The new information is only a time interval on points of the trajectory, which will create a collaborative environment between the controllers.
Many research topics are encompassed by the proposed system. 
First, the definition of the metering points needs an extended analysis from past trajectories. 
Trajectory clustering techniques can be used to determine the flows and the underlying points. 
Thereafter, because of the number of decision variables, the multi-objective context, the graph structure of the trajectories and the use of Monte-Carlo simulations for evaluating the quality of a solution, evolutionary optimization techniques seems good candidates to address the optimization problem. 
Besides, powerful filtering algorithms should be used to keep a consistent model of the actual situation, as the prediction of the sector load depends on such model. 
Finally, because of the stochastic context, a sensitivity analysis should be done on the minimal number of scenarios required to obtain accurate estimates. 
This article describes the theoretical tools that can be used to exactly determine the PDFs. 
Determining the relationship between the rate of convergence of the approximation and the number of required scenarios is left for further work. 
\\
\\
\section{Acknowledgments}
This author Ga{\'e}tan Marceau is funded by the scholarship CIFRE 710/2012 established between Thales Air Systems and INRIA-Saclay, and the scholarship 141138/2011 from the {\it Fonds de Recherche du Qu{\'e}bec - Nature et Technologies}.

\bibliographystyle{IEEEtran}
\bibliography{library}
%

\end{document}